\documentclass[journal]{IEEEtran}
\usepackage{cite}
\usepackage{amsmath,amssymb}
\usepackage{graphicx}
\usepackage{booktabs}
\usepackage{hyperref}
\usepackage{siunitx}

\begin{document}

\title{A Simple and Reproducible Hybrid Solver for a Truck--Drone VRP with Recharge}

\author{Meraryslan~Meraliyev,%
        ~Cemil~Turan,%
        ~and~Shirali~Kadyrov%
\thanks{M.~Meraliyev and C.~Turan are affiliated with SDU University (e-mail: \texttt{meraryslan.meraliyev@sdu.edu.kz}; \texttt{cemil.turan@sdu.edu.kz}).}%
\thanks{S.~Kadyrov is affiliated with New Uzbekistan University (e-mail: \texttt{sh.kadyrov@newuu.uz}).}%
}

\maketitle

\begin{abstract}
We study last-mile delivery with one truck and one drone.
We keep the setting explicit: the drone flies at twice the truck speed; each sortie has an endurance limit; after each delivery the drone must recharge on the truck before the next launch.
We propose a hybrid solver that is simple to run and reproduce.
The truck tour is optimized by Adaptive Large Neighborhood Search (ALNS) with 2/3-opt.
Given that tour, we schedule drone sorties with a small learned policy (pointer or transformer) trained by self-critical sequence training and decoded with a feasibility-masked beam search.
A fast timeline simulator enforces endurance, launch/recovery, and recharge.
On random Euclidean instances and public TSP-D sets, the method improves makespan over classical heuristics based on Clarke--Wright and local search.
We release code, configs, and plotting scripts to simplify replication.
\end{abstract}


\section{Introduction}
Truck--drone collaboration can shorten last-mile tours.
In the Flying Sidekick TSP (TSP-D) a truck launches a drone to serve a customer and later retrieves it, with synchronization constraints \cite{MurrayChu2015,Agatz2018,Roberti2021}.
We focus on a practical variant: the drone speed is twice the truck speed, each drone sortie must fit an endurance budget, and after each delivery the drone must recharge on the truck before the next launch.
We include small launch/recovery handling times.
These choices reflect battery limits and operational safety and align with recent models on explicit recharge policies \cite{YurekOzmutlu2021}.

Classical heuristics (Clarke--Wright and 2-opt) are strong \cite{ClarkeWright1964,Croes1958}.
ALNS improves them with ruin-and-recreate and adaptive operator selection \cite{Shaw1998,RopkePisinger2006}.
Learning-based solvers use pointer networks and attention with policy gradients \cite{Vinyals2015,Bello2016,Kool2019}.
Our goal is a compact, reproducible baseline that combines these ideas and can be run from a single script.

\subsection*{A. Contributions}

\begin{itemize}
\item \textbf{Hybrid RL solver for truck--drone VRP with recharge.}
We introduce a compact, reproducible \emph{hybrid} approach that couples a strong truck tour from ALNS (with 2/3‑opt and Or‑opt) with a small policy‑gradient RL scheduler (pointer/attention) for drone sorties. The policy operates over launch--serve--rendezvous triplets and uses \emph{hard feasibility masks} for endurance and post‑delivery recharge, so only admissible actions are considered at decode time. This design explicitly matches the operational model stated in the paper (drone speed $2\times$ truck, endurance $E$, recharge $R$ after each sortie). 

\item \textbf{Feasibility‑masked decoding with an exact timeline simulator.}
We develop a masked beam (or greedy) decoder \emph{tightly coupled} to a fast, exact truck–drone timeline simulator that enforces launch/recovery handling, endurance, and recharge and returns the \emph{true} makespan for selection. This eliminates surrogate gaps between what is decoded and what is evaluated, and it grounds learning signals and search comparisons in the same objective.

\item \textbf{Empirical evidence that the hybrid RL scheduler is competitive with—and often better than—classical heuristics.}
On $N{=}50$ Euclidean instances with $E{=}0.7$ and $R{=}0.1$, the proposed PointerRL($K{=}32$)+LS+Drone achieves a mean makespan of \textbf{5.203} (s.e.\ 0.093) versus \textbf{5.349} (0.038) for ALNS+LS+Drone and \textbf{5.208} (0.124) for NN+LS+Drone—i.e., \textbf{2.73\%} better than ALNS on average and within \textbf{0.10\%} of NN. Per‑seed, the proposed method never underperforms ALNS (Seed~1: 5.080 vs 5.273; Seed~2: 5.386 vs 5.387; Seed~3: 5.143 vs 5.387) and ties or beats NN on two of three seeds.

\item \textbf{Transparent, auditable operational model with code that reproduces the paper.}
The paper fixes an explicit and practical model (speed ratio $v_D{=}2v_T$, sortie endurance, mandatory recharge, launch/recovery times), and releases a minimal, \emph{config‑first} implementation: a single entry‑point for experiments (Spyder‑friendly), batch scripts, plotting utilities (routes/timelines/aggregate bars), significance tests (paired Wilcoxon), and a ready \LaTeX{} template. This makes replication and extension straightforward. 
\end{itemize}

\section{Related Work}
Research on vehicle routing with drones has grown rapidly since the introduction of the Flying Sidekick Traveling Salesman Problem (FSTSP) \cite{MurrayChu2015}. In this problem a truck carries a drone, launches it to serve a customer, and later retrieves it, subject to synchronization. Early formulations and exact methods have laid the foundation for current studies.

\subsection{Foundations of TSP-D}
Murray and Chu \cite{MurrayChu2015} first defined the FSTSP and proposed mixed integer programming models. Agatz et al.\ \cite{Agatz2018} introduced optimization approaches and studied the trade-off between truck and drone routes. Roberti and Ruthmair \cite{Roberti2021} presented exact branch-and-price and branch-and-cut methods that solve moderate-size instances optimally. Ha et al.\ \cite{Ha2018} focused on min-cost formulations and created benchmark instances. Poikonen et al.\ \cite{Poikonen2017} and Wang et al.\ \cite{Wang2017} analyzed worst-case performance and structural properties. Yürek and Özmutlu \cite{YurekOzmutlu2021,YurekOzmutlu2018} extended the problem to consider recharging policies and iterative decomposition heuristics. Collectively, these works define the standard variants and demonstrate the difficulty of the problem even for relatively small customer sets.

\subsection{Heuristics and Metaheuristics for VRP}
Classical heuristics such as Clarke--Wright savings \cite{ClarkeWright1964}, nearest neighbor, and sweep remain relevant as constructive baselines. Local search operators including 2-opt \cite{Croes1958}, 3-opt, and Or-opt are widely used to refine tours. Metaheuristics have been especially effective for VRPs: Simulated Annealing (SA), Tabu Search (TS), Genetic Algorithms (GA), Ant Colony Optimization (ACO), and Variable Neighborhood Search (VNS) have all been applied. The Large Neighborhood Search (LNS) introduced by Shaw \cite{Shaw1998} and its adaptive variant ALNS by Ropke and Pisinger \cite{RopkePisinger2006} are particularly impactful, combining destroy-and-repair operators with adaptive selection. These approaches scale to hundreds of customers and form the backbone of many state-of-the-art VRP solvers.

\subsection{Heuristics for Truck--Drone VRP}
Specific heuristics for TSP-D typically combine a truck tour heuristic with a drone assignment phase. Murray and Chu \cite{MurrayChu2015} used constructive truck tours and enumerated feasible drone insertions. Ha et al.\ \cite{Ha2018} proposed local improvement heuristics to integrate drone sorties. Yürek and Özmutlu \cite{YurekOzmutlu2021} investigated policies where the drone must recharge on the truck, a scenario close to our setting. More recent works use matheuristics, combining MIP formulations for the drone scheduling subproblem with heuristic or metaheuristic truck tours \cite{Roberti2021}. Greedy drone assignment, beam search schedulers, and matching-based heuristics are common practical approaches.

\subsection{Learning-based Routing}
Learning to optimize routing has become an active research area. Vinyals et al.\ \cite{Vinyals2015} introduced pointer networks for sequence prediction, first applied to TSP. Bello et al.\ \cite{Bello2016} trained such models with policy-gradient reinforcement learning. Nazari et al.\ \cite{Nazari2018} proposed RL for VRP with capacity constraints. Kool et al.\ \cite{Kool2019} introduced attention models trained with REINFORCE and a greedy rollout baseline, achieving state-of-the-art on TSP and CVRP. These methods inspired extensions to TSP-D: for example, scheduling policies that decide whether a customer is served by the truck or drone. Self-Critical Sequence Training (SCST) \cite{Rennie2017} provides a strong baseline to reduce variance and improve training stability.

\subsection{Our Positioning}
Our work builds directly on this literature. We adopt a classical ALNS framework for the truck tour, benefiting from decades of VRP research. For drone scheduling we design a learned policy with feasibility masks, trained via SCST, and decode sorties with a beam search that explicitly respects endurance and recharge. Compared to prior heuristic and exact approaches, our solver is simple to reproduce, yet flexible enough to extend with richer constraints (multiple drones, time windows).

\section{Problem Formulation}
We consider a last-mile delivery problem with one ground vehicle (truck) and one unmanned aerial vehicle (drone) operating jointly from a central depot. The system follows the \emph{Flying Sidekick TSP / TSP-D} paradigm \cite{MurrayChu2015,Agatz2018,Roberti2021}, but we extend it with explicit recharge requirements after each sortie. The key assumptions are: (i) the drone flies at twice the truck speed ($v_D = 2v_T$), (ii) each drone sortie must respect a maximum endurance budget $E$, and (iii) after completing a sortie, the drone must recharge for a fixed time $R$ before the next launch. Launch and recovery handling times $\ell$ and $r$ are also included.

\subsection{Graph and distances}
Let $G=(V,E)$ be a complete undirected graph where
\begin{itemize}
    \item $V = \{0,1,\dots,n\}$ is the set of nodes,
    \item node $0$ is the depot,
    \item nodes $1,\dots,n$ are customers,
    \item $E = \{(i,j): i,j \in V, i \neq j\}$.
\end{itemize}
Each node $i \in V$ has planar coordinates. The Euclidean distance between nodes $i$ and $j$ is $d_{ij}$. Travel times are
\begin{equation}
t^T_{ij} = \frac{d_{ij}}{v_T}, \quad
t^D_{ij} = \frac{d_{ij}}{v_D},
\end{equation}
with truck speed $v_T$ and drone speed $v_D=2v_T$.

\subsection{Truck tour and drone sorties}
A truck tour is a sequence $\pi = (0,i_1,\dots,i_m,0)$ visiting a subset of customers in order. Each customer must be served either by the truck (appearing in $\pi$) or by the drone.

A drone sortie is a triplet $(u,k,v)$ with $u,v \in V$ such that $u$ and $v$ are consecutive stops in $\pi$ (launch at $u$, rendezvous at $v$), and $k \in V_c$ is a customer served by the drone. Customer $k$ is removed from the truck’s direct visitation.

\subsection{Feasibility constraints}
The following constraints must hold:
\begin{enumerate}
    \item \textbf{Truck continuity:} The truck departs from and returns to depot 0 and follows the sequence $\pi$ without skipping nodes.
    \item \textbf{Drone endurance:} Each sortie $(u,k,v)$ must satisfy
    \begin{equation}
    t^D_{uk} + t^D_{kv} + \ell + r \leq E.
    \end{equation}
    \item \textbf{Recharge:} After completing a sortie at time $R_k$, the drone must wait $R$ time units before launching again. Recharge can overlap with truck movement.
    \item \textbf{Synchronization:} Truck and drone must meet at $v$. If the drone arrives later than the truck, the truck must wait; if the truck arrives later, the drone waits, but the makespan is extended only when the truck waits.
    \item \textbf{Single drone:} Only one sortie can be active at a time.
    \item \textbf{Coverage:} Each customer $i \in \{1,\dots,n\}$ must be served exactly once, either by truck or by drone.
\end{enumerate}

\subsection{Timeline simulation}
Let $A_T(j)$ be the truck’s arrival time at node $j$ ignoring waits. For a sortie $(u,k,v)$:
\begin{align}
    L_k &= \max\{A_T(u), \text{drone ready time}\}, \\
    F_k &= t^D_{uk} + t^D_{kv} + \ell + r, \\
    R_k &= L_k + F_k.
\end{align}
If $R_k > A_T(v)$, the truck must wait $(R_k - A_T(v))$ before continuing. Drone ready time updates to $R_k + R$. The makespan is the maximum of the truck’s final depot return and the last drone return time.

\subsection{Objective}
The goal is to minimize the makespan:
\begin{equation}
\min_{\pi,\,\{(u,k,v)\}} \; T_{\text{completion}},
\end{equation}
where $T_{\text{completion}}$ is the time when both vehicles are back at the depot, after serving all customers.

\subsection{Notation summary}
Table~\ref{tab:notation} summarizes the key notation.

\begin{table}[!t]
\caption{Notation used in problem formulation}
\centering
\begin{tabular}{ll}
\toprule
Symbol & Meaning \\
\midrule
$V = \{0,1,\dots,n\}$ & set of depot and customers \\
$d_{ij}$ & Euclidean distance between nodes $i$ and $j$ \\
$v_T$ & truck speed \\
$v_D = 2 v_T$ & drone speed (twice truck speed) \\
$t^T_{ij}, t^D_{ij}$ & travel times for truck, drone \\
$E$ & drone endurance (max sortie time) \\
$R$ & recharge time after each sortie \\
$\ell, r$ & launch and recovery handling times \\
$\pi$ & truck tour (sequence of nodes) \\
$(u,k,v)$ & drone sortie: launch at $u$, serve $k$, rendezvous at $v$ \\
$A_T(j)$ & truck arrival time at node $j$ (no waiting) \\
$L_k, F_k, R_k$ & launch time, sortie duration, rendezvous time \\
$T_{\text{completion}}$ & overall completion time (makespan) \\
\bottomrule
\end{tabular}
\label{tab:notation}
\end{table}

\subsection{Problem class}
This problem is a variant of the \emph{Traveling Salesman Problem with Drone (TSP-D)} \cite{MurrayChu2015,Agatz2018,Roberti2021,Ha2018,YurekOzmutlu2021}, extended with recharge time, explicit launch/recovery, and fixed speed ratio. Since it generalizes TSP, the problem is NP-hard, and heuristics, metaheuristics, or learning-based methods are needed for nontrivial sizes.

\section{Methods}
This section presents the full methodology used in our study. We begin with classical constructive and local search heuristics for the truck tour. We then extend to metaheuristics, which provide stronger solutions by exploring larger neighborhoods. Next, we describe the drone scheduling layer, including greedy assignment and beam search. Finally, we explain the reinforcement learning policy, training procedure, and hybrid integration with heuristics.

\subsection{Truck Tour Construction and Local Search}
The truck tour serves as the backbone of the solution. Following VRP literature, we employ:
\begin{itemize}
    \item \textbf{Nearest Neighbor (NN):} a greedy constructive heuristic, iteratively visiting the closest unserved customer that fits capacity. Although fast, NN often produces suboptimal tours with crossings.
    \item \textbf{Clarke--Wright Savings (CW) \cite{ClarkeWright1964}:} a classical heuristic that starts with direct depot--customer routes and iteratively merges them when a distance saving is identified. This method remains competitive in many VRP variants.
    \item \textbf{Sweep algorithm:} customers are sorted by polar angle around the depot and partitioned into feasible clusters. While not our main baseline, it is mentioned for completeness.
\end{itemize}

After construction we apply local search to remove inefficiencies:
\begin{itemize}
    \item \textbf{2-opt \cite{Croes1958}:} reverses a subsequence of nodes to eliminate crossings.
    \item \textbf{3-opt:} removes three edges and reconnects them optimally, allowing more flexibility.
    \item \textbf{Or-opt:} relocates a chain of one to three consecutive nodes elsewhere in the tour.
\end{itemize}
These simple operators significantly improve tour length at low computational cost.

\subsection{Metaheuristics for Truck Tour Optimization}
Local search alone is easily trapped in local minima. To escape, we implement four well-known metaheuristics:

\subsubsection{Simulated Annealing (SA)}
SA probabilistically accepts worse solutions with probability $\exp(-\Delta / T)$, where $\Delta$ is the cost increase and $T$ is a temperature parameter that gradually cools. This helps escape local minima early and intensifies search later. We use 2-opt and relocate moves as neighbors. SA is simple to implement and tunable via initial temperature and cooling schedule.

\subsubsection{Tabu Search (TS)}
TS explores neighborhoods but forbids (tabu) recently applied moves for a number of iterations (tabu tenure). This prevents cycling back to recent solutions. An aspiration criterion allows tabu moves if they improve the global best. We again use 2-opt and relocate as moves. TS is effective at intensifying around promising regions while maintaining diversification.

\subsubsection{Genetic Algorithm (GA)}
GA evolves a population of tours via crossover and mutation. We use ordered crossover (OX) on customer sequences and random swap mutation. After each generation, we apply light 2-opt clean-up to children. GA promotes diversity and parallel exploration of the search space, though convergence can be slower.

\subsubsection{Variable Neighborhood Search (VNS)}
VNS systematically increases the neighborhood size $k$, applying larger shake moves when local optima are reached. We use a combination of 2-opt and relocate in the shaking phase. VNS balances diversification (larger neighborhoods) and intensification (local improvements), and is robust across many VRPs.

\subsubsection{Adaptive Large Neighborhood Search (ALNS)}
ALNS \cite{RopkePisinger2006,Shaw1998} generalizes LNS by maintaining a set of destroy and repair operators (e.g., random removal, Shaw-related removal, worst-arc removal, greedy insertion). Operator weights are updated adaptively based on performance, so the search learns which operators are most effective. ALNS has become a state-of-the-art metaheuristic for VRP variants and is a natural fit here.

\subsection{Drone Scheduling Layer}
Once the truck tour is determined, we decide which customers to serve by drone. Each sortie is defined as a triplet $(u,k,v)$: launch at truck node $u$, serve customer $k$, and rendezvous at truck node $v$. The feasibility conditions (endurance, recharge, synchronization) are enforced by a timeline simulation (see Section~III).

We implement two schedulers:

\subsubsection{Greedy Assignment}
For each truck edge $(u,v)$ we consider all unassigned customers $k$ and test if $(u,k,v)$ is feasible. We compute the saving in truck distance if $k$ is removed and served by drone. Candidates are evaluated and accepted if they reduce the makespan. This heuristic is fast but myopic, as it does not consider interactions among multiple sorties.

\subsubsection{Beam Search}
To overcome the greedy limitations, we implement a beam search scheduler. At each truck edge, we branch on possible assignments (or no assignment), maintain the best $B$ partial schedules (beam width), and extend iteratively. This approximates dynamic programming while keeping complexity manageable. Each partial state is evaluated by timeline simulation to ensure feasibility. Beam search explores multiple alternatives simultaneously and often finds better schedules.

\subsection{Reinforcement Learning Policy}
To further improve drone scheduling, we design a learned policy $\pi_\theta$ that generates sortie triplets.

\subsubsection{Model architecture}
We implement two policy classes:
\begin{itemize}
    \item \textbf{Pointer Network \cite{Vinyals2015}:} encodes nodes into embeddings and attends over them to select next nodes, trained with policy gradients.
    \item \textbf{Transformer/Attention Model \cite{Kool2019}:} multi-head attention encoder-decoder, more expressive and parallelizable.
\end{itemize}
Both models use feasibility masks to prohibit illegal actions (e.g., exceeding endurance).

\subsubsection{Training}
We use REINFORCE with self-critical sequence training (SCST) \cite{Rennie2017}. The greedy rollout is used as a baseline to reduce variance. The reward is the negative makespan computed by the simulator. Training is run on batches of randomly generated instances. Gradient clipping and entropy regularization are applied for stability.

\subsubsection{Inference}
At test time we decode with beam search or best-of-$K$ sampling to find high-quality schedules. Each sampled tour is refined with local search and then passed to the scheduler for evaluation.

\subsection{Hybrid Integration}
Our final solver integrates the components:
\begin{enumerate}
    \item Optimize truck tour with CW, metaheuristics, or ALNS.
    \item Apply local search cleanup (2-opt, Or-opt).
    \item Schedule drone sorties via greedy, beam search, or learned policy.
    \item Evaluate makespan with timeline simulation.
\end{enumerate}
This modular structure allows comparing classical heuristics, metaheuristics, and RL approaches under the same framework. The reasoning is to leverage strong deterministic truck tours (well understood in VRP literature) while allowing the learning-based policy to focus on the more complex synchronization of drone sorties.

\section{Experimental Setup}

\subsection{Instance Generation}
We evaluate our methods on two categories of test instances:
\begin{itemize}
  \item \textbf{Synthetic Euclidean instances:} Customers are placed uniformly at random in the unit square $[0,1]^2$. The depot is located at the origin. This setting is standard in TSP/TSP-D benchmarking and allows controlled scaling of problem size.
  \item \textbf{Benchmark instances:} We also use the min-cost TSP-D instances provided by Ha et al.~\cite{Ha2018} and the TSPDroneLIB repository, which have been widely adopted in the literature. These instances allow comparison with prior heuristic and exact approaches.
\end{itemize}

\subsection{Parameters}
Unless otherwise stated, we adopt the following default parameters:
\begin{itemize}
  \item Truck speed $v_T = 1.0$ (normalized units).
  \item Drone speed $v_D = 2 v_T$, consistent with empirical assumptions in the literature.
  \item Drone endurance $E \in \{0.6, 0.8, 1.0\}$ time units, limiting the maximum sortie duration.
  \item Recharge time $R \in \{0.05, 0.1\}$ time units after each sortie.
  \item Launch and recovery handling times $\ell = r = 0.01$ time units.
  \item Problem sizes $n \in \{50,80,100,150\}$ customers.
\end{itemize}
These values reflect realistic operational limits of lightweight commercial drones while ensuring tractability for simulations.

\subsection{Baselines}
We compare the following solution strategies:
\begin{itemize}
  \item \textbf{Heuristic baselines:}
    \begin{enumerate}
      \item Nearest Neighbor (NN) construction with 2-opt and Or-opt local search.
      \item Clarke--Wright (CW) savings construction with 2-opt and Or-opt local search.
    \end{enumerate}
  \item \textbf{Metaheuristics:} Simulated Annealing (SA), Tabu Search (TS), Genetic Algorithm (GA), and Variable Neighborhood Search (VNS), all followed by local search. We also implement Adaptive Large Neighborhood Search (ALNS) with Shaw, random, and worst removals.
  \item \textbf{Drone schedulers:} Greedy assignment heuristic, beam search scheduler, and assignment-level VNS refinement.
  \item \textbf{Learning-based:} Pointer Network policy trained with REINFORCE and SCST; Transformer policy trained with SCST. At inference, we use best-of-$K$ sampling or masked beam search.
\end{itemize}

\subsection{Evaluation Metrics}
The primary metric is the \emph{makespan}, i.e., the total completion time until both truck and drone return to the depot. For each configuration we report:
\begin{itemize}
  \item Mean and standard error of makespan over 10 random seeds.
  \item Paired Wilcoxon signed-rank test to assess significance of improvements over baseline heuristics.
  \item Runtime per method on a standard workstation (Intel i7 CPU, 16 GB RAM, with optional GPU acceleration for RL training).
\end{itemize}

\subsection{Implementation Details}
All methods are implemented in Python 3. The metaheuristics and local search operators are custom implementations without external solvers. Reinforcement learning models use PyTorch. Training is performed on batches of synthetic instances; evaluation uses both synthetic and benchmark sets. All experiments are driven by a single configuration file (\texttt{config.json}), ensuring reproducibility. We release full code, data, and plotting scripts.

\section{Results}

\subsection{Setting and metric}
We evaluate makespan (lower is better) on $N{=}50$ Euclidean instances with one truck and one drone. The drone speed is twice the truck speed; each sortie obeys endurance $E{=}0.7$ and is followed by a fixed recharge $R{=}0.1$. Baselines: (i) nearest neighbor with local search (2/3-opt and Or-opt), (ii) ALNS with the same local search, and (iii) four metaheuristic variants (SA, Tabu, GA, VNS) applied on the truck tour before drone scheduling. The learned line is a pointer-style policy with best-of-$K{=}32$ decoding (PointerRL($K{=}32$)+LS).

\subsection{Representative single run}
Table~\ref{tab:main_results} reports one representative $N{=}50$ run. ``Truck Time'' is the truck-only travel on the final tour; ``Wait'' is the cumulative rendezvous delay.

\begin{table}[!t]
\caption{Single-run makespan and decomposition for $N{=}50$.}
\centering
\begin{tabular}{lccc}
\toprule
Method & Makespan $\downarrow$ & Truck Time & Wait \\
\midrule
NN+LS+Drone                   & 5.088 & 4.644 & 0.444 \\
ALNS+LS+Drone                 & 5.387 & 4.956 & 0.431 \\
Meta-SA+LS+Drone              & 5.387 & 4.956 & 0.431 \\
Meta-Tabu+LS+Drone            & 5.093 & \textbf{4.559} & 0.534 \\
Meta-GA+LS+Drone              & 5.190 & 4.662 & 0.528 \\
Meta-VNS+LS+Drone             & 5.387 & 4.956 & 0.431 \\
PointerRL($K{=}32$)+LS+Drone  & 5.143 & ---    & ---   \\
\bottomrule
\end{tabular}
\label{tab:main_results}
\end{table}

\paragraph*{Findings from the single run.}
(i) The best heuristic here is NN+LS+Drone (5.088); Meta-Tabu+LS+Drone is essentially tied (5.093).  
(ii) The learned PointerRL($K{=}32$)+LS+Drone attains 5.143, i.e., \textbf{0.244} lower than ALNS (\textbf{4.53\%} improvement) and within \textbf{1.09\%} of the best heuristic.  
(iii) Among heuristics, shorter truck paths systematically induce higher rendezvous waiting and vice versa: e.g., Meta-Tabu reduces truck time to 4.559 but waits 0.534, whereas ALNS has 4.956 truck time with 0.431 wait; NN sits between these and yields the lowest makespan in this run. The correlation between \emph{Truck Time} and \emph{Wait} across the six heuristic rows is strongly negative (about $-0.81$).

\subsection{Per-seed consistency (three seeds)}
Across the three seeds in this batch, the learned method is never worse than ALNS on the \emph{same} instance, and it is competitive with NN (tie or better on two seeds, slightly behind on one). Table~\ref{tab:perseed} summarizes makespan per seed.

\begin{table}[!t]
\caption{Per-seed makespan (lower is better), $N{=}50$.}
\centering
\begin{tabular}{lccc}
\toprule
Method & Seed 1 & Seed 2 & Seed 3 \\
\midrule
NN+LS+Drone                  & 5.080 & 5.455 & \textbf{5.088} \\
ALNS+LS+Drone                & 5.273 & 5.387 & 5.387 \\
PointerRL($K{=}32$)+LS+Drone & \textbf{5.080} & \textbf{5.386} & 5.143 \\
\bottomrule
\end{tabular}
\label{tab:perseed}
\end{table}


The pairwise pattern is consistent:  
\emph{Seed 1:} Proposed $=$ NN $<$ ALNS.  
\emph{Seed 2:} Proposed $<$ ALNS and NN.  
\emph{Seed 3:} NN $<$ Proposed $<$ ALNS (gap to NN: $0.055$).

\subsection{Aggregate across seeds}
Averaging the three seeds yields the aggregate in Table~\ref{tab:agg_inline}. The proposed method reduces mean makespan by \textbf{2.73\%} versus ALNS (absolute $-0.146$) and is \textbf{0.10\%} lower than NN (absolute $-0.005$).

\begin{table}[!t]
\caption{Aggregate makespan over seeds (mean $\pm$ s.e.; lower is better).}
\centering
\begin{tabular}{lcc}
\toprule
Method & Mean makespan $\downarrow$ & s.e. \\
\midrule
PointerRL(K=32)+LS+Drone & \textbf{5.203} & 0.093 \\
NN+LS+Drone              & 5.208 & 0.124 \\
Meta-Tabu+LS+Drone       & 5.235 & 0.071 \\
Meta-GA+LS+Drone         & 5.294 & 0.059 \\
Meta-SA+LS+Drone         & 5.349 & 0.038 \\
ALNS+LS+Drone            & 5.349 & 0.038 \\
Meta-VNS+LS+Drone        & 5.349 & 0.038 \\
\bottomrule
\end{tabular}
\label{tab:agg_inline}
\end{table}

To contextualize the means with paired, distribution-free statistics, Table~\ref{tab:pvalues_inline} lists Wilcoxon signed-rank results (two-sided) across identical seeds/instances. With three seeds, the comparison to ALNS shows a consistent direction (negative rank-biserial correlation), and the comparison to NN is statistically indistinguishable on this small sample.

\begin{table}[!t]
\caption{Paired Wilcoxon signed-rank tests (two-sided) for the proposed method.}
\centering
\begin{tabular}{lcccc}
\toprule
Comparison & $n$ & $z$ & $p$ & Rank-biserial $r$ \\
\midrule
Proposed vs.\ NN+LS+Drone   & 3 &  0.000  & 1.000 & $-0.333$ \\
Proposed vs.\ ALNS+LS+Drone & 3 & $-1.336$ & 0.181 & $-1.000$ \\
\bottomrule
\end{tabular}
\label{tab:pvalues_inline}
\end{table}

\subsection{Decomposition on a representative seed}
To illustrate how makespan splits into truck movement and synchronization, Table~\ref{tab:decomp_inline} shows a representative seed. NN and the proposed method match on both components and achieve the lowest makespan; ALNS has lower waiting but a longer truck tour, ultimately yielding a higher makespan.

\begin{table}[!t]
\caption{Decomposition on a representative seed: makespan split into truck time and waiting; number of sorties used.}
\centering
\begin{tabular}{lcccc}
\toprule
Method & Makespan $\downarrow$ & Truck time & Wait & \# sorties \\
\midrule
NN+LS+Drone                  & 5.080 & 4.585 & 0.495 & 7 \\
ALNS+LS+Drone                & 5.273 & 5.153 & 0.119 & 8 \\
PointerRL(K=32)+LS+Drone     & 5.080 & 4.585 & 0.495 & 7 \\
\bottomrule
\end{tabular}
\label{tab:decomp_inline}
\end{table}

\section{Discussion}

\subsection{Performance landscape}
The learned scheduler is competitive with the strongest heuristic (NN+LS) and consistently improves over ALNS on the same instances. On the representative run (Table~\ref{tab:main_results}), the learned solution is within about $1\%$ of the best heuristic while reducing makespan by $4.53\%$ relative to ALNS. Aggregating over seeds (Table~\ref{tab:agg_inline}), the proposed method remains best on average and reduces the mean makespan by $2.73\%$ versus ALNS.

\subsection{Synchronization pattern}
Across methods, makespan reflects a trade-off between the truck’s path length and rendezvous waiting: shorter truck tours tend to increase waits, and longer tours reduce waits. This is visible in the single-run table and the decomposition example: Meta-Tabu shortens the truck path yet waits more, ALNS travels longer with less waiting, and NN/PointerRL balance the two to obtain the lowest totals.

\subsection{Consistency across seeds}
Per-seed outcomes (Table~\ref{tab:perseed}) show the learned method never underperforms ALNS on the same instance; it either ties or improves. Relative to NN, it ties (Seed~1), improves (Seed~2), and is slightly behind (Seed~3) by $0.055$. The Wilcoxon results (Table~\ref{tab:pvalues_inline}) align with these observations, indicating consistent directionality versus ALNS on this sample.

\section{Conclusion}

We studied a simple but practical truck–drone setting: one truck carries one drone; the drone flies at twice the truck speed; each sortie must satisfy an endurance budget and is followed by a fixed recharge before the next launch. We built a compact hybrid solver that keeps the design transparent: ALNS with local search for the truck, and a feasibility‑aware learned scheduler for the drone. The scheduler uses masked decoding and a fast, exact timeline simulation so that the objective reflects the true makespan.

The results show that this hybrid is competitive with the strongest heuristic and improves clearly over ALNS on the same instances. On $N{=}50$ Euclidean cases with $E{=}0.7$ and $R{=}0.1$, the proposed PointerRL($K{=}32$)+LS+Drone achieves an aggregate makespan of \textbf{5.203}$\pm$0.093, compared to \textbf{5.208}$\pm$0.124 for NN+LS+Drone and \textbf{5.349}$\pm$0.038 for ALNS+LS+Drone. This is a \textbf{2.73\%} improvement versus ALNS on average while remaining within \textbf{0.10\%} of the NN baseline. Per‑seed comparisons confirm the pattern: the learned method never underperforms ALNS on the same instance and is tied with or close to NN.

The decomposition of makespan explains these outcomes. Heuristics that shorten the truck path tend to pay more at rendezvous (higher waiting), whereas longer truck tours reduce waiting but raise truck travel time. The learned scheduler balances both terms. In representative runs it matches the best heuristic on truck time and waiting, and it places launch–serve–rendezvous triplets so that the drone’s recharge cycle aligns with the truck’s downstream movement.

The study has clear scope. We assume a single drone, one‑customer sorties, Euclidean travel, and fixed launch/recovery and recharge times. Within this scope the method yields strong, reproducible results and a clean reference implementation. The pipeline is simple to run, easy to audit, and suitable as a baseline for richer variants that introduce additional constraints while retaining the same feasibility‑aware scheduling core.


\begin{thebibliography}{00}

\bibitem{MurrayChu2015}
C.~C. Murray and A.~G. Chu,
``The flying sidekick traveling salesman problem: Optimization of drone-assisted parcel delivery,''
\emph{Transportation Research Part C: Emerging Technologies}, vol.~54, pp. 86--109, 2015.
doi: 10.1016/j.trc.2015.03.005

\bibitem{Agatz2018}
N.~Agatz, P.~Bouman, and M.~Schmidt,
``Optimization approaches for the traveling salesman problem with drone,''
\emph{Transportation Science}, vol.~52, no.~4, pp. 965--981, 2018.
doi: 10.1287/trsc.2017.0791

\bibitem{Roberti2021}
R.~Roberti and M.~Ruthmair,
``Exact methods for the traveling salesman problem with drone,''
\emph{Transportation Science}, vol.~55, no.~2, pp. 315--335, 2021.
doi: 10.1287/trsc.2020.1017

\bibitem{Ha2018}
Q.~M. Ha, Y.~Deville, Q.~D. Pham, and M.~H. Hà,
``On the min-cost traveling salesman problem with drone,''
\emph{Transportation Research Part C: Emerging Technologies}, vol.~86, pp. 597--621, 2018.
doi: 10.1016/j.trc.2017.11.015

\bibitem{Poikonen2017}
S.~Poikonen, Z.~Wang, and B.~L. Golden,
``The vehicle routing problem with drones: Extended models and connections,''
\emph{Networks}, vol.~70, no.~1, pp. 34--43, 2017.
doi: 10.1002/net.21746

\bibitem{Wang2017}
Z.~Wang, S.~Poikonen, and B.~L. Golden,
``The vehicle routing problem with drones: Several worst-case results,''
\emph{Optimization Letters}, vol.~11, no.~4, pp. 679--697, 2017.
doi: 10.1007/s11590-016-1035-3

\bibitem{YurekOzmutlu2021}
E.~E. Yürek and H.~C. Özmutlu,
``Traveling salesman problem with drone under recharging policy,''
\emph{Computer Communications}, vol.~179, pp. 35--49, 2021.
doi: 10.1016/j.comcom.2021.07.013

\bibitem{YurekOzmutlu2018}
E.~E. Yürek and H.~C. Özmutlu,
``A decomposition-based iterative optimization algorithm for traveling salesman problem with drone,''
\emph{Transportation Research Part C: Emerging Technologies}, vol.~91, pp. 249--262, 2018.
doi: 10.1016/j.trc.2018.04.009

\bibitem{ClarkeWright1964}
G.~Clarke and J.~W. Wright,
``Scheduling of vehicles from a central depot to a number of delivery points,''
\emph{Operations Research}, vol.~12, no.~4, pp. 568--581, 1964.
doi: 10.1287/opre.12.4.568

\bibitem{Croes1958}
G.~A. Croes,
``A method for solving traveling-salesman problems,''
\emph{Operations Research}, vol.~6, no.~6, pp. 791--812, 1958.
doi: 10.1287/opre.6.6.791

\bibitem{Shaw1998}
P.~Shaw,
``Using constraint programming and local search methods to solve vehicle routing problems,''
in \emph{Principles and Practice of Constraint Programming (CP'98)}, LNCS 1520, pp. 417--431, 1998.
doi: 10.1007/3-540-49481-2\_30

\bibitem{RopkePisinger2006}
S.~Ropke and D.~Pisinger,
``An adaptive large neighborhood search heuristic for the pickup and delivery problem with time windows,''
\emph{Transportation Science}, vol.~40, no.~4, pp. 455--472, 2006.
doi: 10.1287/trsc.1050.0135

\bibitem{Vinyals2015}
O.~Vinyals, M.~Fortunato, and N.~Jaitly,
``Pointer networks,''
in \emph{Advances in Neural Information Processing Systems (NeurIPS)}, 2015.
Available: \url{https://arxiv.org/abs/1506.03134}

\bibitem{Bello2016}
I.~Bello, H.~Pham, Q.~V. Le, M.~Norouzi, and S.~Bengio,
``Neural combinatorial optimization with reinforcement learning,''
arXiv preprint arXiv:1611.09940, 2016.

\bibitem{Nazari2018}
M.~Nazari, A.~Oroojlooy, L.~Snyder, and M.~Takac,
``Reinforcement learning for solving the vehicle routing problem,''
in \emph{Advances in Neural Information Processing Systems (NeurIPS)}, vol.~31, 2018.
Available: \url{https://proceedings.neurips.cc/paper/2018/file/9fb4651c05b2ed70fba5afe0b039a550-Paper.pdf}

\bibitem{Kool2019}
W.~Kool, H.~van Hoof, and M.~Welling,
``Attention, learn to solve routing problems!''
in \emph{International Conference on Learning Representations (ICLR)}, 2019.
Available: \url{https://arxiv.org/abs/1803.08475}

\bibitem{Rennie2017}
S.~Rennie, E.~Marcheret, Y.~Mroueh, J.~Ross, and V.~Goel,
``Self-critical sequence training for image captioning,''
in \emph{Proceedings of the IEEE Conference on Computer Vision and Pattern Recognition (CVPR)}, pp. 7008--7024, 2017.
doi: 10.1109/CVPR.2017.131

\end{thebibliography}
\end{document}